\pdfoutput=1

\documentclass[11pt]{article}

\usepackage[final]{acl}
\usepackage{amsmath}
\usepackage{graphicx} 

\usepackage{times}
\usepackage{latexsym}

\usepackage[T1]{fontenc}

\usepackage[utf8]{inputenc}

\usepackage{microtype}

\usepackage{inconsolata}
\usepackage{booktabs}
\usepackage{float}      
\usepackage{placeins}   


\author{AS Aravinthakshan \\ Manipal Institute of Technology \\ Perssonify
        \And
        Laven Srivastava \\ Perssonify
        \And
        Harsh Nandwani \\ Perssonify
        \AND
        \normalfont\texttt{research@perssonify.com}}

\title{Human Agreement and Return Association Are Not Interchangeable Criteria}

\begin{document}
\maketitle

\begin{abstract}
Financial NLP has a standard workflow: validate a sentiment tool against human labels, then trust it to extract market signal. This assumes the two evaluations measure the same thing. We test that assumption in a setting where both can be measured at once: a corpus of securities class actions (2002--2025) linking 70,500 X\footnote{formerly Twitter} messages to abnormal stock returns, with a single-annotator human labelled gold sample. Running five instruments (VADER, Loughran--McDonald, FinBERT, Twitter-RoBERTa, and an LLM annotator) through one identical pipeline, we find that the relationship between construct and predictive validity depends on the sampling convention and score representation. Under conventional method-specific sampling, human agreement aligns more closely with graded same-day associations than with one-day leads. On a fixed-\(n\) panel, however, agreement has similar graded rank correlations at both horizons, while the coarse ordering remains weak. Benchmark agreement therefore establishes semantic validity but does not by itself determine predictive rankings. In a conversation that is 17.6\% spam, message volume predicts neither market damage nor settlement size.

\end{abstract}

\section{Introduction}
\label{sec:intro}

Sentiment instruments are validated in natural language processing the way
classifiers are generally validated: by agreement with human annotation on
a labeled sample. In finance, they are used for a different purpose, to
capture a signal that is associated with, and ideally precedes, movements
in asset prices. These two criteria are routinely treated as though the
first implied the second. An instrument that scores well on a sentiment
benchmark is assumed to be the better instrument for market signal
extraction. That assumption is rarely tested, because the two evaluations
are normally conducted on different corpora, human labeled benchmarks
carry no market outcomes and market datasets carry no human labels.

This paper tests the assumption directly: We construct a corpus in which
the same messages carry both. Securities class actions provide the
linkage: court complaints supply event anchors and litigation metadata,
the defendant's price history supplies abnormal returns, and the
surrounding social media conversation supplies text that we annotate both
automatically, with five instruments, and manually, on a gold sample.
All instruments use the same aggregation and testing procedures. We compare both method-specific samples and a shared set of days, since the days retained can also affect the results.

The domain is deliberately adverse; corrective disclosures and filings
generate genuine investor discussion, but also plaintiff firm
solicitation, repeated news headlines, and automated promotion. The
solicitation is structural, the PSLRA requires the first filing plaintiff
to publish notice inviting other investors to seek lead plaintiff
appointment, so every case mechanically produces a burst of law firm press
releases \citep{choi2024business}. Headline repetition is a known feature
of financial news \citep{tetlock2011}, and bots piggyback the cashtags of
newsworthy firms to promote unrelated securities \citep{cresci2019}. Spam
is 17.6\% of our corpus, and solicitation and spam together account for
25{,}166 messages. Counting is therefore a poor measure by construction: a
count absorbs all of these sources equally. We ask three questions.



\begin{itemize}
\item \textbf{RQ1.} Does agreement with human sentiment judgments predict
      the strength of an instrument's association with abnormal returns?
\item \textbf{RQ2.} If agreement and association diverge, at what horizon
      do they diverge, and what are the instruments responding to?
\item \textbf{RQ3.} Which conclusions about litigation discourse are
      invariant to the choice of instrument, and which are not?
\end{itemize}

Our contributions are as follows.

\begin{enumerate}
\item \textbf{A dual validity corpus, released.} We release 66{,}890
event anchored messages across 845 securities class actions
(2002--2025), carrying five instrument sentiment labels and multi axis
LLM annotation, anchored by a 400 message human gold standard with its
codebook and agreement statistics; to our knowledge, the first public
financial social media corpus in which the same messages carry human
annotation and linked market outcomes, so both validities can be
measured through one pipeline.\footnote{\url{https://bit.ly/4iSUf6d}}

\item \textbf{A sampling dependent relationship between the two validities. } Under conventional method-specific sampling, human agreement aligns more closely with graded same-day associations than with one-day leads. On the fixed-\(n\) panel, the graded ordering is similar at both horizons, whereas the coarse ordering remains weak. The relationship between construct and predictive validity therefore depends on horizon, representation, and the treatment of zero-score days.

\item \textbf{Mechanism, and instability of instrument rankings.}
      Under conventional method-specific sampling, the instrument with the largest one-day point estimate is pretrained on financial news; news accounts carry the strongest attributed signal (\(\rho=-0.261\)), and the local-projection placebo fails at negative horizons.
\item \textbf{Domain result and informative nulls.} Content is
      informative where volume is not, despite systematic pollution;
      message volume predicts neither the depth of the price decline nor
      dollar settlement size.
\end{enumerate}

\section{Related Work}
\label{sec:related}

\paragraph{Content and volume in financial discourse.}
The distinction between what market participants say and how much they
post originates in the message board literature: \citet{tumarkin2001}
find little return information in posting activity, while
\citet{antweiler2004} show that message \emph{content} carries signal
where volume mainly predicts volatility, and \citet{daschen2007} build
the first purpose made sentiment classifier for stock talk. Media
pessimism predicts market activity \citep{tetlock2007}, firm level
language predicts fundamentals \citep{tetlock2008}, and the content of
crowd-sourced investment opinions predicts returns and earnings
surprises \citep{chen2014}. Message volume is one member of a family of attention measures that includes search intensity \citep{da2011} and the attention driven trading it induces \citep{barber2008}. We treat this content over volume
regularity as established and use it as a domain check; the paper's
claim concerns how the instruments that measure content should be
evaluated.

\paragraph{Sentiment instruments and their benchmark evaluation.}
The instruments we compare span the standard toolkit: a social media
lexicon \citep{hutto2014}, a finance dictionary \citep{loughran2011},
and transformers pretrained on financial news \citep{araci2019} and on
tweets \citep{barbieri2020}. Such instruments are standardly evaluated
by agreement with human annotation, with Financial PhraseBank
\citep{malo2014} the reference benchmark; recent suites extend the same
paradigm to financial large language models
\citep{xie2023pixiu,xie2024finben}, making the question of what
agreement predicts downstream more consequential, not less.
\citet{loughran2016} argue that measurement choices dominate downstream
conclusions; our results give that argument a sharp form: the choice
between two validated instruments changes which temporal conclusions
are recoverable.

\paragraph{Intrinsic versus extrinsic evaluation.}
The distinction we draw between agreement with human labels and
association with outcomes is the psychometric distinction between
construct and predictive validity \citep{cronbach1955}, and it has an
NLP precedent: intrinsic evaluations of word representations fail to
predict extrinsic task performance \citep{chiu2016}, and
measurement-theoretic critiques argue that NLP systems routinely
operationalise constructs without testing what the operationalisation
measures \citep{jacobs2021}. To our knowledge the two evaluations have
not previously been conducted on the same financial messages with
linked market outcomes, which is what permits the horizon localised
comparison in Section~\ref{sec:divergence}.

\paragraph{LLMs in financial text.}
Large language models extract return relevant signal from headlines
\citep{lopezlira2023} and outperform transformer and dictionary
sentiment in trading settings \citep{kirtac2024}; LLM annotation can
match or exceed crowd workers \citep{gilardi2023}. Two caveats shape
our design: LLM sentiment can embed look ahead information from the
pretraining window \citep{glasserman2023}, motivating the contamination
analysis of Section~\ref{sec:robustness}, with chronologically
consistent training the proposed remedy \citep{he2025}.

\paragraph{Social media, returns, and securities litigation.}
Relations between social media mood and market movement are established
\citep{bollen2011,sprenger2014,ranco2015}, as is the link from investor
disagreement to trading volume \citep{cookson2020}; class action event
studies measure shareholder wealth effects, litigation risk, and
reputational penalties without social media text
\citep{gande2009,kim2012,karpoff2008}. Our setting differs in being
event anchored by court filings, carrying eventual legal outcomes, and
evaluating the measurement instruments themselves rather than any one
instrument's signal. The press is a documented fraud detection channel
\citep{miller2006,dyck2010}, consistent with our attribution result
that news accounts carry the strongest price relevant signal and with
the news arrival reading of the leading component.
\section{Corpus and Annotation}
\label{sec:data}

\subsection{Corpus construction}
The corpus contains 845 securities class actions in three cohorts
(Table~\ref{tab:corpus}): prospective 2024 and 2025 cohorts and a
retrospective archive of resolved cases from 2002--2021. Court filings
supply company identity, class period bounds, corrective disclosure dates,
allegation summaries, defendants, and litigation metadata. Filings are
ingested from PDF and converted to structured records. High value
fields (company name, ticker, class period bounds, defendants, and
disclosure dates) receive an independent extraction pass, and
disagreements are manually reconciled. These dates define the
message retrieval and event windows.

\begin{table}[t]
\centering\small
\begin{tabular}{@{}lrrrrr@{}}
\toprule
Cohort & Cases & Ticker & Stock & Social & Msgs. \\
\midrule
2024    & 204 & 204 & 162 & 56  & 15{,}984 \\
2025    & 182 & 180 & 154 & 51  & 10{,}178 \\
Archive & 459 & 305 & 180 & 161 & 44{,}338 \\
\midrule
Total   & 845 & 689 & 496 & 268 & 70{,}500 \\
\bottomrule
\end{tabular}
\caption{Corpus by cohort. ``Social'' counts cases with collected
messages; ``Msgs.'' counts annotated messages.}
\label{tab:corpus}
\end{table}

\subsection{Composition of the conversation}
\label{sec:composition}

The label distributions in this subsection are produced by the LLM
instrument described in Section~\ref{sec:instruments}. The relevance,
topic, and polarity taxonomies are closed sets fixed in the annotation
prompt, which is released with the code and data.

Of 70{,}500 messages, 62.0\% are relevant, 20.3\% tangential, and 17.6\%
spam. The largest topics are news reporting (13{,}880), law firm
solicitation (12{,}901), automated spam (12{,}265), legal procedure
(9{,}658), equity analysis (8{,}055), and direct fraud allegations
(7{,}835). Polarity is 41.0\% negative and
7.2\% positive. Relevance and polarity agree substantially with human
annotation ($\kappa = 0.746$ and $0.741$; Section~\ref{sec:construct}),
while the topic taxonomy is not separately validated and is therefore used
descriptively only.

A count based attention measure aggregates informative reporting, investor
reaction, solicitation, and automation into a single number, motivating
the comparison in Section~\ref{sec:predictive}.

\subsection{Event and market linkage}
Messages are assigned to a pre class period baseline, the alleged class
period, the corrective disclosure window, and post disclosure and
post filing windows, retaining text, timestamp, impressions, and likes
where available. For company $i$ on day $t$, let $R_{i,t}$ denote the realised return and
$R_{m,t}$ the market return. A market model
\begin{equation}
R_{i,t} = \alpha_i + \beta_i R_{m,t} + \epsilon_{i,t}
\end{equation}
is estimated over the 120 trading days ending ten days before the class
period begins, so the parameters are fitted on pre event data only. The
\emph{abnormal return} is the part of the day's return the market does not
explain,
\begin{equation}
\mathit{AR}_{i,t} = R_{i,t} - \big(\hat{\alpha}_i + \hat{\beta}_i R_{m,t}\big),
\end{equation}
and the \emph{cumulative abnormal return} over an event window $W$ is its
sum, $\mathit{CAR}_i = \sum_{t \in W} \mathit{AR}_{i,t}$.

\subsection{Scoring instruments}
\label{sec:instruments}
Five instruments score message sentiment: VADER, a social media lexicon
\citep{hutto2014}; Loughran--McDonald, a finance dictionary
\citep{loughran2011}; FinBERT and Twitter-RoBERTa, transformers pretrained
on financial news and on tweets \citep{araci2019,barbieri2020}; and Claude
Haiku, an LLM annotator. The instruments produce different outputs, so we derive coarse and graded negativity scores, with higher values indicating more negative sentiment. The \emph{coarse} score discards magnitude. Each message is assigned to one
of three classes and mapped to $\{+1,0,-1\}$ with $+1$ negative, using the
predicted class for the transformers and the LLM, and the sign of the score
for the lexicons. The \emph{graded} score keeps magnitude, oriented so that larger is more
negative: the negated VADER compound score, the Loughran--McDonald
net-negative word ratio, $p(\text{neg}) - p(\text{pos})$ for the
transformers, and the negated $-2$ to $+2$ intensity label for the LLM.

Both scores are averaged over the messages for each case day. All instruments use the same aggregation and tests, but the conventional samples differ because zero-score days are excluded separately for each instrument. Section~\ref{sec:instability} repeats the comparison on a shared set of days.

\subsection{Construct validity}
\label{sec:construct}
We assess agreement with human judgment on a gold set of 400 messages,
labelled by a single annotator blind to all model outputs and presented in
randomised order. Messages were drawn by stratified random sampling across
predicted relevance $\times$ polarity $\times$ cohort (26 strata);
inverse probability weighted accuracy (0.861 polarity, 0.860 relevance)
matches the unweighted estimates, so stratification does not drive the
reported agreement.

Table~\ref{tab:construct} reports polarity agreement for all five
instruments. We report Cohen's $\kappa$ \citep{cohen1960} alongside
accuracy because the corpus is 41.0\% negative and 7.2\% positive, so an
instrument that predicts the majority class attains substantial accuracy
while carrying no information. FinBERT is precisely this case: accuracy
0.538 with $\kappa = 0.079$. Under conventional benchmarks
\citep{landis1977} only the LLM instrument reaches substantial agreement;
Twitter-RoBERTa is moderate, Loughran--McDonald fair, and FinBERT
indistinguishable from chance. VADER's accuracy is below the majority class
rate.

The LLM instrument's polarity errors concentrate at the negative neutral
boundary and are directionally one sided: it labels human neutral messages
negative in 49 of 56 disagreements, so LLM derived negativity is if
anything slightly inflated. On relevance it achieves 0.860 accuracy,
0.797 macro-F1, and $\kappa = 0.746$. This evaluation measures agreement
with the intended linguistic constructs; it makes no claim about
association with returns, which is the subject of the next two sections.

\begin{table}[t]
\centering\small
\setlength{\tabcolsep}{4pt}
\begin{tabular}{@{}lrrr@{}}
\toprule
Instrument & Accuracy & Macro-F1 & $\kappa$ \\
\midrule
Claude Haiku       & 0.860 [0.825, 0.893] & 0.878 & 0.741 \\
RoBERTa    & 0.733 [0.693, 0.778] & 0.594 & 0.449 \\
LM & 0.553 [0.508, 0.598] & 0.446 & 0.210 \\
FinBERT            & 0.538 [0.490, 0.588] & 0.379 & 0.079 \\
VADER              & 0.385 [0.338, 0.438] & 0.373 & 0.171 \\
\bottomrule
\end{tabular}
\caption{Polarity agreement with human gold labels on the same 400
messages. Brackets give 95\% bootstrap intervals from 1{,}000 resamples
\citep{efron1993}. LM denotes Loughran--McDonald; RoBERTa denotes
Twitter-RoBERTa.}
\label{tab:construct}
\end{table}
\section{Empirical Setup}
\label{sec:setup}

We report Spearman $\rho$ for monotonic associations, which is robust to
the heavy tails of daily returns. Daily analyses pool case days;
cross sectional analyses collapse each case to one observation.
Significance markers are $^{***}p<0.001$, $^{**}p<0.01$, $^{*}p<0.05$,
$^{\dagger}p<0.1$. New baseline families use Benjamini--Hochberg FDR
adjustment \citep{benjamini1995}. Because the daily panel is large,
statistical significance is easily attained; we therefore compare
instruments on effect size and treat $p$-values as evidence only that an
association is nonzero.

Timing proceeds from descriptive to conditional tests. Lead lag
correlations compare sentiment on $t-\ell$ with $AR_t$. Order two Granger
tests ask whether lagged sentiment improves prediction beyond return
history \citep{granger1969}. A distributed lag model,
\begin{equation}
AR_{i,t} = \sum_{k=0}^{3}\beta_k s_{i,t-k} + \eta_i + u_{i,t},
\end{equation}
uses case fixed effects and case clustered standard errors
\citep{petersen2009}. Local projections \citep{jorda2005} estimate
$AR_{i,t+h}$ on $s_{i,t}$ for $h \in [-5,5]$, with negative horizons
serving as a placebo. These tests establish predictive precedence, not
structural causation. Between-instrument differences in $ \rho $ are not directly tested; comparisons of their magnitudes are therefore descriptive point-estimate comparisons, and significance for one instrument but not another does not establish a significant difference between them.

The panel contains 179 cases and 103{,}542 case days.\footnote{The panel
was re fetched because the original price series were unavailable. Of 273
archive tickers, 74 no longer resolve. Published values reproduce within
$|\Delta\rho| \le 0.002$, with identical signs and significance ordering.
All cross method comparisons use the refreshed same data panel.} The
conventional correlation sample drops days on which a method emits zero,
so $n$ differs by method; Section~\ref{sec:instability} reports a
common day intersection in which $n$ is held fixed.

\section{Predictive Validity: Content versus Counting}
\label{sec:predictive}

At case level, attention does not track damage. In the 2025 cohort, total
message volume is uncorrelated with the worst single-day abnormal return
($\rho = -0.083$, $p = 0.60$, $n = 43$) and with CAR ($\rho = +0.003$,
$p = 0.99$, $n = 36$).

\begin{figure*}[t]
\centering
\includegraphics[width=.92\textwidth]{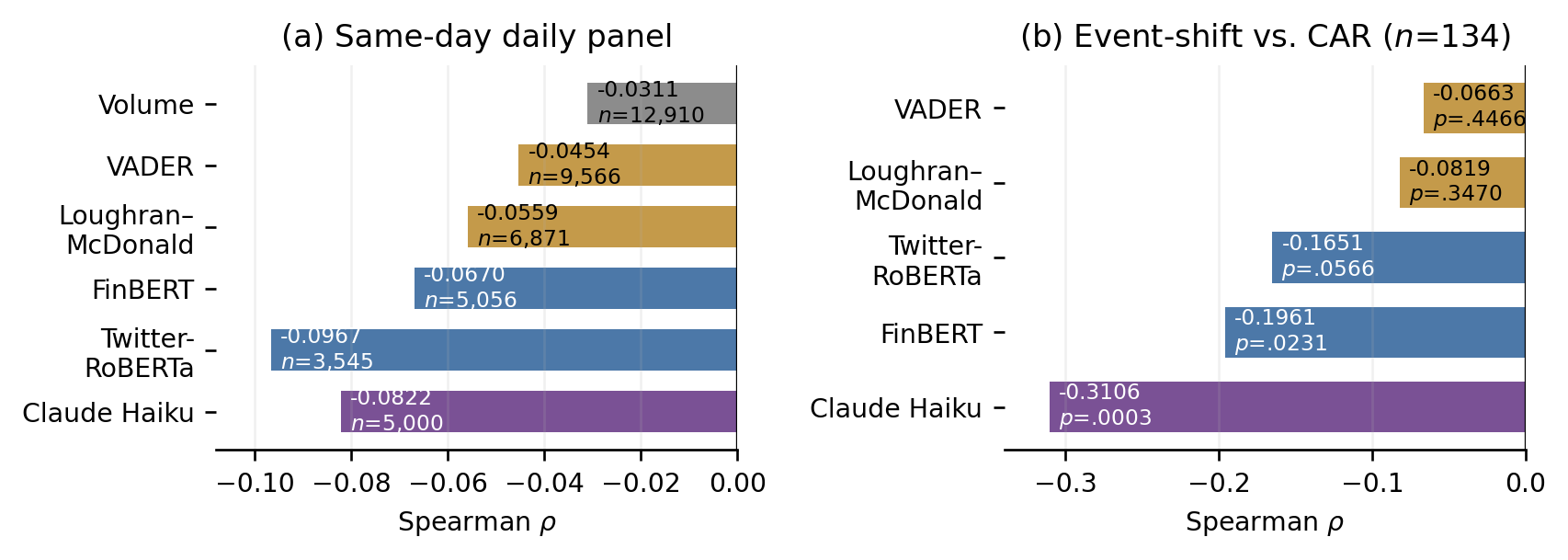}
\caption{Content over volume across scoring instruments. Panel (a) shows the strongest same day representation in each method family; $n$ differs because the conventional analysis drops method specific zero days. Panel (b) uses the identical 134 cases for event minus baseline sentiment shift versus CAR. Bars report exact $\rho$ values without treating any instrument as the contribution.}
\label{fig:content}
\end{figure*}

The daily panel establishes the comparison precisely
(Table~\ref{tab:sameday}, summarised in Figure~\ref{fig:content}). Tweet volume reaches $\rho = -0.0311$. Every
sentiment family contains a representation with a larger association, and
the strongest instruments exceed volume by a factor of two to three.
Loughran--McDonald coarse ($0.80\times$) is the single exception, so the
claim applies to method families rather than to every threshold choice.
Impressions, the natural reach based count, are not significant at all.

Two features of this table matter for what follows. First, the strongest
same day instrument is Twitter RoBERTa, not the LLM, so the
content over counting result does not depend on the LLM annotation.
Second, the $p$-values are not comparable across rows: RoBERTa graded
attains $p = 5.27\times10^{-18}$ on 12{,}910 case days while FinBERT
coarse attains $p = 1.87\times10^{-6}$ on 5{,}056, and the difference is
largely sample size. We compare the $\rho$ column throughout.

\begin{table}[t]
\centering\small
\setlength{\tabcolsep}{2.5pt}
\begin{tabular}{@{}lrrrr@{}}
\toprule
Metric & $\rho$ & $p$ & $n$ & $\times$vol. \\
\midrule

\multicolumn{5}{@{}l}{\textit{Counts}}\\
Tweet volume   & $-0.0311^{***}$ & $4.14{\times}10^{-4}$ & 12{,}910 & $1.00\times$ \\
Impressions    & $-0.0317^{\dagger}$ & $0.0634$ & 3{,}439 & $1.02\times$ \\
Negative count & $-0.0380^{*}$ & $0.0135$ & 4{,}219 & $1.22\times$ \\

\midrule

    \multicolumn{5}{@{}l}{\textit{Lexicons}}\\
VADER c. & $-0.0454^{***}$ & $8.86{\times}10^{-6}$ & 9{,}566 & $1.46\times$ \\
VADER g. & $-0.0454^{***}$ & $3.78{\times}10^{-6}$ & 10{,}339 & $1.46\times$ \\
LM c.    & $-0.0249^{*}$ & $0.0414$ & 6{,}713 & $0.80\times$ \\
LM g.    & $-0.0559^{***}$ & $3.60{\times}10^{-6}$ & 6{,}871 & $1.80\times$ \\

\midrule

\multicolumn{5}{@{}l}{\textit{Transformers}}\\
FinBERT c. & $-0.0670^{***}$ & $1.87{\times}10^{-6}$ & 5{,}056 & $2.16\times$ \\
FinBERT g. & $-0.0526^{***}$ & $2.29{\times}10^{-9}$ & 12{,}910 & $1.69\times$ \\
RoBERTa c. & $-0.0967^{***}$ & $7.91{\times}10^{-9}$ & 3{,}545 & $3.11\times$ \\
RoBERTa g. & $-0.0760^{***}$ & $5.27{\times}10^{-18}$ & 12{,}910 & $2.45\times$ \\

\midrule

\multicolumn{5}{@{}l}{\textit{LLM instrument}}\\
Claude c.  & $-0.0822^{***}$ & $5.77{\times}10^{-9}$ & 5{,}000 & $2.65\times$ \\
Claude g.  & $-0.0821^{***}$ & $6.12{\times}10^{-9}$ & 5{,}004 & $2.64\times$ \\

\bottomrule
\end{tabular}
\caption{Same day correlation with the daily abnormal return. LM denotes
Loughran--McDonald; RoBERTa denotes Twitter-RoBERTa; c.\ and g.\ denote
coarse and graded scores. $\times$vol. is $|\rho|$ relative to tweet volume.}
\label{tab:sameday}
\end{table}

\section{Temporal Structure}
\label{sec:timing}

Table~\ref{tab:leadlag} reports lead lag correlations. Every instrument
shows a negative and significant same day association, so the
contemporaneous result of Section~\ref{sec:predictive} is not
instrument specific. Under the conventional method-specific nonzero-day samples, the one-day results differ across instruments. FinBERT coarse has the largest \(\ell=1\) point estimate (\(-0.0466\), \(p_{\mathrm{FDR}}=0.003\)); FinBERT graded and Twitter-RoBERTa graded also survive FDR, whereas VADER and Loughran--McDonald do not. At \(\ell=2\), Claude graded and Twitter-RoBERTa graded are significant at the unadjusted 5\% level. These comparisons describe point estimates and do not establish statistically significant differences between instruments.

\begin{table}[t]
\centering\small
\setlength{\tabcolsep}{2pt}
\begin{tabular}{@{}lrrrr@{}}
\toprule
Method & $\rho_0$ & $\rho_1$ & $q_1$ & $\rho_2$ \\
\midrule
VADER c.   & $-0.0454^{***}$ & $-0.0136$                & 0.286          & $+0.0025$ \\
VADER g.   & $-0.0454^{***}$ & $-0.0193^{*}$            & 0.100          & $-0.0034$ \\
LM c.      & $-0.0249^{*}$   & $-0.0057$                & 0.810          & $+0.0168$ \\
LM g.      & $-0.0559^{***}$ & $-0.0263^{*}$            & 0.067          & $-0.0057$ \\
FinBERT c. & $-0.0670^{***}$ & $\mathbf{-0.0466^{***}}$ & \textbf{0.003} & $+0.0010$ \\
FinBERT g. & $-0.0526^{***}$ & $-0.0292^{***}$          & 0.003          & $-0.0022$ \\
RoBERTa c. & $-0.0967^{***}$ & $-0.0058$                & 0.879          & $+0.0010$ \\
RoBERTa g. & $-0.0760^{***}$ & $-0.0272^{**}$           & 0.006          & $-0.0202^{*}$ \\
Claude c.  & $-0.0822^{***}$ & $-0.0227$                & ---            & $-0.0022$ \\
Claude g.  & $-0.0821^{***}$ & $-0.0376^{**}$           & ---            & $-0.0373^{**}$ \\
\bottomrule
\end{tabular}
\caption{Lead lag correlations, all messages; c.\ and g.\ denote
coarse and graded scores; $n$ per row as in Table~\ref{tab:sameday}.
$q$ values are FDR adjusted across the baseline family; the pre existing
LLM reference is not included in that correction.}
\label{tab:leadlag}
\end{table}

Conditional tests give the same picture. Four baseline specifications
survive FDR on forward Granger tests: Loughran--McDonald coarse/relevant
($F = 4.388$, $q = 0.034$), Loughran--McDonald graded/all ($F = 4.056$,
$q = 0.040$), FinBERT coarse/relevant ($F = 3.907$, $q = 0.044$), and
FinBERT graded/all ($F = 4.316$, $q = 0.034$). The LLM reference also
passes ($F = 5.435$, $p = 0.0051$ coarse/all). Twitter-RoBERTa never
survives, despite having the strongest same day association. Same day
distributed lag coefficients are negative and significant for almost every
instrument; their magnitudes are not comparable across instruments because
the underlying signals differ in scale. Full results appear in
Appendix~\ref{app:granger}.



\section{Where the Two Validities Diverge}
\label{sec:divergence}

Sections~\ref{sec:construct} and~\ref{sec:timing} measure two different
properties of the same five instruments on the same corpus: agreement with
human judgment, and association with abnormal returns. Table~\ref{tab:joint}
places them side by side. This is the paper's central comparison, and it
is not visible from either table in isolation.

\begin{table*}[t]
\centering\small
\begin{tabular}{@{}lrrrrrrr@{}}
\toprule
& \multicolumn{3}{c}{Construct validity} & \multicolumn{4}{c}{Predictive validity ($|\rho|$)} \\
\cmidrule(lr){2-4}\cmidrule(lr){5-8}
Instrument & $\kappa$ & Accuracy & Macro-F1 & coarse $\rho_0$ & coarse $\rho_1$ & graded $\rho_0$ & graded $\rho_1$ \\
\midrule
Claude Haiku       & 0.741 & 0.860 & 0.878 & 0.0822 & 0.0227 & \textbf{0.0821} & \textbf{0.0376} \\
Twitter-RoBERTa    & 0.449 & 0.733 & 0.594 & \textbf{0.0967} & 0.0058 & 0.0760 & 0.0272 \\
Loughran--McDonald & 0.210 & 0.553 & 0.446 & 0.0249 & 0.0057 & 0.0559 & 0.0263 \\
VADER              & 0.171 & 0.385 & 0.373 & 0.0454 & 0.0136 & 0.0454 & 0.0193 \\
FinBERT            & 0.079 & 0.538 & 0.379 & 0.0670 & \textbf{0.0466} & 0.0526 & 0.0292 \\
\midrule
\multicolumn{4}{@{}l}{Rank correlation with $\kappa$}
  & $+0.50$ & $-0.30$ & $+0.90^{\dagger}$ & $+0.40$ \\
\multicolumn{4}{@{}l}{Rank correlation with accuracy}
  & $+0.60$ & $-0.10$ & $+1.00^{*}$ & $+0.70$ \\
\bottomrule
\end{tabular}
\caption{Construct validity from Table~\ref{tab:construct} joined to predictive validity from Tables~\ref{tab:sameday} and~\ref{tab:leadlag}, ordered by \(\kappa\). On conventional method-specific nonzero-day samples, human-agreement rankings align most closely with graded same-day association and less consistently with the one-day lead. Rank correlations are Spearman correlations over five instruments, with exact two-sided \(p\)-values from all 120 permutations. These comparisons are descriptive, and no cell survives adjustment for the 12 comparisons in this table. The corresponding fixed-\(n\) comparison is reported in Table~\ref{tab:common} and Section~\ref{sec:instability}.}
\label{tab:joint}
\end{table*}

Conventional samples show contemporaneous alignment. For the graded representation in Table~\ref{tab:joint}, the ordering by human agreement closely matches the ordering by same-day association: the rank correlation is \(+0.90\) with \(\kappa\) and \(+1.00\) with accuracy. Thus, under this representation and the conventional method-specific sampling rule, benchmark agreement is informative about contemporaneous association.


The one-day relationship is weaker and depends on the score representation. For graded scores, the rank correlations are $+0.40$ with $\kappa$ and $+0.70$ with accuracy; for coarse scores, they are $-0.30$ and $-0.10$. Under conventional sampling, FinBERT has the strongest coarse one-day association ($\rho_1 = -0.0466$) despite its low human agreement; Claude's corresponding value is $-0.0227$. Section~\ref{sec:instability} shows how these rankings change when all instruments are evaluated on the same case-day set.

\paragraph{Interpretation.}
Table~\ref{tab:leadlag} shows a horizon contrast under the conventional, method-specific samples, but Table~\ref{tab:common} shows that this contrast is not invariant to sample construction or score representation. On the fixed-$n$ graded panel, agreement has the same rank correlation with association at $\ell = 0$ and $\ell = 1$; on the coarse panel, both relationships are weak. We therefore make the narrower claim that benchmark agreement does not, by itself, determine predictive rankings, and that comparisons should be reported with the horizon, representation, and zero-day inclusion rule explicitly stated.

\subsection{Instrument rankings are unstable}
\label{sec:instability}

The conventional analysis drops case days on which a given method emits
zero, so $n$ differs across instruments in
Tables~\ref{tab:sameday} and~\ref{tab:leadlag} and the comparison is not
made on identical data. Table~\ref{tab:common} restricts to case days on
which every instrument emits a nonzero score, holding $n$ fixed.

\begin{table}[t]
\centering\small
\setlength{\tabcolsep}{4pt}
\begin{tabular}{@{}lrrr@{}}
\toprule
Method & $\rho_0$ & $\rho_1$ & $\rho_2$ \\
\midrule
\multicolumn{4}{@{}l}{\textit{Coarse, all messages} ($n=1{,}518$)}\\
Claude  & $-0.0851^{***}$          & $-0.0323$           & $-0.0101$ \\
VADER   & $-0.0068$                & $-0.0165$           & $+0.0300$ \\
LM      & $-0.0460^{\dagger}$      & $-0.0501^{\dagger}$ & $-0.0333$ \\
FinBERT & $\mathbf{-0.1259^{***}}$ & $-0.0315$           & $-0.0116$ \\
RoBERTa & $-0.0667^{**}$           & $-0.0073$           & $-0.0030$ \\
\midrule
\multicolumn{4}{@{}l}{\textit{Graded, all messages} ($n=3{,}890$)}\\
Claude  & $-0.0696^{***}$     & $-0.0438^{**}$           & $-0.0467^{**}$ \\
VADER   & $-0.0288^{\dagger}$ & $-0.0173$                & $+0.0081$ \\
LM      & $-0.0411^{*}$       & $-0.0339^{*}$            & $-0.0035$ \\
FinBERT & $-0.0385^{*}$       & $-0.0297^{\dagger}$      & $+0.0215$ \\
RoBERTa & $-0.0723^{***}$     & $\mathbf{-0.0561^{***}}$ & $-0.0189$ \\
\bottomrule
\end{tabular}
\caption{Common-case-day intersection, all messages.
Relevant-message panels appear in Appendix~\ref{app:common}.}
\label{tab:common}
\end{table}

Applying the Table~\ref{tab:joint} rank comparison to the fixed-$n$ panels gives correlations with $\kappa$ of +0.00 and +0.10 for coarse scores at $\ell = 0$ and 1, and +0.80 at both horizons for graded scores. The corresponding values for accuracy and macro-F1 are +0.40 and +0.20 for coarse scores, and +0.90 at both graded horizons. The exact two-sided permutation $p$-values are 1.000, 0.950, 0.133, and 0.133 for $\kappa$, and 0.517, 0.783, 0.083, and 0.083 for accuracy and macro-F1, respectively. None of these correlations survives BH adjustment within the 12 fixed-\(n\) construct--predictive comparisons.

Three observations follow. First, the same day ranking reverses: on the
coarse common days FinBERT is strongest ($-0.1259$), where the
conventional sample placed Twitter-RoBERTa first. Second, VADER is
indistinguishable from zero on common days, so its apparent advantage in
Table~\ref{tab:sameday} was substantially a matter of which days it scored.
Third, several sentiment measures remain associated with returns on the shared days, although the strongest instrument changes. Table~\ref{tab:common} does not compare sentiment with volume on those same days, so it does not establish that the content-over-counting result is unchanged. Claims about the strongest instrument therefore need to specify the sample used.

\subsection{Divergence at the case level}
\label{sec:caselevel}

A cross sectional test reproduces the divergence in a different form. For
each case we compute the event minus baseline change in mean negativity
and correlate it with the event window CAR (Table~\ref{tab:caselevel}).
The LLM instrument attains $\rho = -0.3106$ ($p = 0.0003$, $n = 134$),
several times any daily panel effect. No baseline survives FDR
correction: FinBERT reaches $-0.1961$ ($p = 0.0231$) but
$q = 0.185$, and Twitter-RoBERTa $-0.1651$ ($p = 0.0566$). The case level shift is a contemporaneous contrast between two windows
rather than a lead, so the $\ell = 0$ pattern of Table~\ref{tab:joint}
would predict the best agreeing instrument to perform well here, which it
does. We do not press that reading further, for two reasons. With
$n = 134$ and five instruments the comparison is underpowered, so the
ordering among the baselines is not resolvable; and the result is
sensitive to how the case level measure is constructed. Substituting mean
window negativity for negative share, on a reconstruction of the panel,
raises Loughran--McDonald and Twitter-RoBERTa to the same range as the LLM
instrument. The robust content of this table is that a case level
contrast exists and is several times larger than any daily panel effect;
the identity of the strongest instrument at this level is not established.
Restricting to relevant messages reduces the sample to $n = 50$ and
renders every instrument nonsignificant.

\begin{table}[t]
\centering\small
\begin{tabular}{@{}lrrr@{}}
\toprule
Instrument & $\rho$ & $p$ & $q$ \\
\midrule
Claude Haiku       & $-0.3106^{***}$ & 0.0003 & --- \\
FinBERT            & $-0.1961^{*}$ & 0.0231 & 0.185 \\
Twitter-RoBERTa    & $-0.1651^{\dagger}$ & 0.0566 & 0.226 \\
Loughran--McDonald & $-0.0819$ & 0.3470 & 0.596 \\
VADER              & $-0.0663$ & 0.4466 & 0.596 \\
\bottomrule
\end{tabular}
\caption{Event-minus-baseline change in the share of negative messages versus CAR, using the same 134 cases for all instruments and including all messages.}
\label{tab:caselevel}
\end{table}

\section{What the Leading Signal Responds To}
\label{sec:mechanism}

Section~\ref{sec:divergence} shows that the relationship between construct validity and the one-day lead depends on the sampling convention and score representation. Two pieces of evidence bear on what the observed leading associations reflect.


\paragraph{The largest conventional-sample lead is produced by an instrument trained on news.}
FinBERT is pretrained on financial news and evaluated on sentence-level news sentiment. On tweets about securities fraud, it agrees with human polarity at close to chance, yet its coarse score has the largest one-day point estimate under conventional method-specific sampling. The natural reading is that it detects the vocabulary of adverse financial news rather than the polarity of investor expression, and that this vocabulary arrives shortly before the price adjustment completes.

\paragraph{News accounts carry the strongest signal.} We classify accounts from topic mix, cross-case activity, and spam fraction. News accounts have the strongest association ($\rho = -0.261$, $p < 0.001$), followed by cross-case broadcasters ($-0.161$), other multi-case accounts ($-0.115$), and retail accounts ($-0.083$). The associations for bot/spam accounts ($+0.064$) and law firm solicitation ($+0.040$) are not significant. The strongest observed signal therefore comes from news accounts, while these two sources of additional volume show no detectable association with returns.


\section{Robustness}
\label{sec:robustness}

\paragraph{Look ahead contamination.}
The archive cohort (2002--2021) lies inside the LLM's pretraining
window, so its results could in principle reflect recognition of
publicised cases rather than message level reading
\citep{glasserman2023}. The baselines provide a control: the two
dictionaries have no training window at all, and the two transformers
were pretrained without these outcome labels. Contamination would
therefore show up as an LLM advantage that is larger in the archive
than in the 2024--2025 cohorts.
On the daily panel it is not. Across 1{,}000 case level bootstrap
resamples, the archive minus recent difference in the LLM's advantage
is $+0.0091$ (graded, same day; 95\% CI $[-0.027, +0.046]$), $-0.0013$
(graded, one day lag; CI $[-0.047, +0.052]$), $+0.0061$ and $+0.0017$
(coarse). Every interval includes zero. The LLM's own association is also similar across cohorts ($-0.081$ in the archive and $-0.090$ in the recent cohorts). These results do not reveal a clear difference between cohorts, but they do not rule out contamination. The LLM's own association is stable across cohorts
($-0.081$ archive, $-0.090$ recent). We retain the archive for the
daily analyses.
The case level measure of Section~\ref{sec:caselevel} is less settled:
split by cohort (80 archive, 47 recent), the LLM's advantage is larger
in the archive the direction contamination predicts. The subsample
is too small to treat as evidence either way, so the paper rests its
claims on the daily panel and reports the case level result as
corroborative (decomposition in Appendix~\ref{app:contamination}).

\paragraph{Relevance filtering.}
Restricting to relevant messages shifts same day correlations by at
most $0.0164$; the central result is present on all messages. Since the
relevance labels come from the LLM instrument, this removes a
circularity concern (Appendix~\ref{app:filtering}).


\paragraph{Attention and legal materiality.}
Among 133 resolved cases with settlements, total message volume is
uncorrelated with settlement amount ($\rho = -0.041$, $p = 0.64$), as
are relevant message ($-0.062$) and negative message ($-0.093$) volume;
median settlements do not differ across volume tertiles ($H = 0.83$,
$p = 0.66$). Public attention is orthogonal to legal materiality, as it
was to price damage depth.

\section{Discussion and Conclusion}
\label{sec:discussion}

We introduce a corpus linking 845 securities class actions to 70,500 event-anchored messages, pairing a 400-message single-annotator human reference sample with abnormal-return series linked to the messages. Evaluating five instruments through one pipeline, we find that the relationship between human agreement and return association is sensitive to sampling and representation. On conventional method-specific samples, human agreement closely orders graded contemporaneous associations but not one-day leads; on the fixed-\(n\) panel, graded rankings are similar at both horizons, while coarse rankings remain weak.

Agreement with human labels is therefore evidence of semantic validity, not a sufficient basis for selecting an instrument for predictive use. Construct and predictive validity should be reported together with the horizon, score representation, and zero-day inclusion rule.

The account-level results and the failed placebo are consistent with a leading signal that partly reflects news arrival. The daily-panel comparison finds no clear archive-recent difference in the LLM's relative advantage, but does not rule out contamination. The domain
result stands: in a conversation that is 17.6\% spam, content informs
where counting does not, and volume predicts neither the depth of the
price decline nor the size of the eventual settlement. The contribution is
an evaluation design and a measurement result, not a classifier.

\section*{Limitations}

Human validation covers polarity and relevance but not intensity, emotion,
severity, or topic, and it reflects agreement with a single careful
annotator rather than inter annotator reliability; no human to human
agreement figure is available, so the $\kappa$ values in
Table~\ref{tab:construct} have no measured ceiling.

The rank comparisons are based on five instruments and are therefore descriptive. Under conventional method-specific sampling, human agreement aligns more closely with graded same-day associations than with one-day leads. We therefore conclude only that benchmark agreement does not by itself determine predictive rankings and that the sampling convention and score representation must accompany such comparisons. Widening the instrument battery is a natural next step.

FinBERT's low agreement is partly attributable to domain shift: it was
tuned on Financial PhraseBank sentences rather than on tweets. We regard
this as consistent with, rather than an alternative to, our reading, since
practitioners apply it off the shelf to exactly this kind of text; but it
means that $\kappa = 0.079$ should be read as agreement in deployment
rather than as an intrinsic property of the model.

Account types are behavioural rather than profile derived.
Relevance filtered case level windows become sparse ($n$ falls from 134 to
50). Reach weighted variants have been evaluated only for the LLM
instrument. Timing tests establish predictive precedence rather than
structural causality or tradeability, and the local projection placebo
fails at negative horizons. Effect sizes throughout are small.

The corpus is specific to securities-class-action discourse, where solicitation, repeated headlines, and cashtag piggybacking are unusually prevalent. Although the evaluation design is portable, whether the observed relationships generalize to ordinary stock discussion requires cross-domain validation.

In addition, 74 of 273 archive tickers no longer resolve through the original price API. If unresolved symbols disproportionately represent delisted firms, their absence from the refreshed price panel may truncate the adverse-return tail and attenuate estimated negative associations. The reported return relationships should therefore be interpreted subject to this potential survivorship bias.

\section*{Ethics Statement}

The corpus consists of public X/Twitter posts collected via the platform
API. No author profile lookups were performed; account categories are
derived solely from posting behaviour, and all attribution results are
reported at the aggregate class level. No personally identifying
information is included in released artifacts. Court filings are public
records. In accordance with the platform's developer terms, the released
artifact contains message identifiers and our annotations rather than
message text, together with a script for retrieving text from the
platform.

\section*{Data and Code Availability}

The provided archive contains the classification prompts and
configuration, the 400 message gold annotation file, message identifiers
with all model and human labels, per instrument daily sentiment series,
the computed abnormal return panel, and analysis code sufficient to
reproduce every table. The abnormal return panel is included directly
because 74 of 273 archive tickers no longer resolve through the original
price API.

\section*{Acknowledgements}


We extend our gratitude to Perssonify LLC. for their continued support of our work, and allocating resources that made this work possible. In particular we thank Stefan Persson for his valuable inputs and guidance.

\bibliography{references}

@article{cohen1960,
  author = {Cohen, Jacob},
  title = {A Coefficient of Agreement for Nominal Scales},
  journal = {Educational and Psychological Measurement},
  year = {1960}, volume = {20}, number = {1}, pages = {37--46}}

@article{landis1977,
  author = {Landis, J. Richard and Koch, Gary G.},
  title = {The Measurement of Observer Agreement for Categorical Data},
  journal = {Biometrics},
  year = {1977}, volume = {33}, number = {1}, pages = {159--174}}

@book{efron1993,
  author = {Efron, Bradley and Tibshirani, Robert J.},
  title = {An Introduction to the Bootstrap},
  publisher = {Chapman \& Hall}, year = {1993}}

@article{benjamini1995,
  author = {Benjamini, Yoav and Hochberg, Yosef},
  title = {Controlling the False Discovery Rate: A Practical and Powerful Approach to Multiple Testing},
  journal = {Journal of the Royal Statistical Society: Series B},
  year = {1995}, volume = {57}, number = {1}, pages = {289--300}}

@article{granger1969,
  author = {Granger, Clive W. J.},
  title = {Investigating Causal Relations by Econometric Models and Cross-Spectral Methods},
  journal = {Econometrica},
  year = {1969}, volume = {37}, number = {3}, pages = {424--438}}

@article{jorda2005,
  author = {Jord{\`a}, {\`O}scar},
  title = {Estimation and Inference of Impulse Responses by Local Projections},
  journal = {American Economic Review},
  year = {2005}, volume = {95}, number = {1}, pages = {161--182}}

@article{petersen2009,
  author = {Petersen, Mitchell A.},
  title = {Estimating Standard Errors in Finance Panel Data Sets: Comparing Approaches},
  journal = {Review of Financial Studies},
  year = {2009}, volume = {22}, number = {1}, pages = {435--480}}

@article{antweiler2004,
  author  = {Antweiler, Werner and Frank, Murray Z.},
  title   = {Is All That Talk Just Noise? {T}he Information Content of Internet Stock Message Boards},
  journal = {The Journal of Finance},
  year    = {2004}, volume = {59}, number = {3}, pages = {1259--1294}}

@article{tumarkin2001,
  author  = {Tumarkin, Robert and Whitelaw, Robert F.},
  title   = {News or Noise? {I}nternet Postings and Stock Prices},
  journal = {Financial Analysts Journal},
  year    = {2001}, volume = {57}, number = {3}, pages = {41--51}}

@article{daschen2007,
  author  = {Das, Sanjiv R. and Chen, Mike Y.},
  title   = {Yahoo! for {A}mazon: Sentiment Extraction from Small Talk on the Web},
  journal = {Management Science},
  year    = {2007}, volume = {53}, number = {9}, pages = {1375--1388}}

@article{tetlock2007,
  author  = {Tetlock, Paul C.},
  title   = {Giving Content to Investor Sentiment: The Role of Media in the Stock Market},
  journal = {The Journal of Finance},
  year    = {2007}, volume = {62}, number = {3}, pages = {1139--1168}}

@article{tetlock2008,
  author  = {Tetlock, Paul C. and Saar-Tsechansky, Maytal and Macskassy, Sofus},
  title   = {More Than Words: Quantifying Language to Measure Firms' Fundamentals},
  journal = {The Journal of Finance},
  year    = {2008}, volume = {63}, number = {3}, pages = {1437--1467}}

@article{chen2014,
  author  = {Chen, Hailiang and De, Prabuddha and Hu, Yu (Jeffrey) and Hwang, Byoung-Hyoun},
  title   = {Wisdom of Crowds: The Value of Stock Opinions Transmitted Through Social Media},
  journal = {Review of Financial Studies},
  year    = {2014}, volume = {27}, number = {5}, pages = {1367--1403}}

@article{da2011,
  author  = {Da, Zhi and Engelberg, Joseph and Gao, Pengjie},
  title   = {In Search of Attention},
  journal = {The Journal of Finance},
  year    = {2011}, volume = {66}, number = {5}, pages = {1461--1499}}

@article{barber2008,
  author  = {Barber, Brad M. and Odean, Terrance},
  title   = {All That Glitters: The Effect of Attention and News on the Buying Behavior of Individual and Institutional Investors},
  journal = {Review of Financial Studies},
  year    = {2008}, volume = {21}, number = {2}, pages = {785--818}}

@article{loughran2011,
  author  = {Loughran, Tim and McDonald, Bill},
  title   = {When Is a Liability Not a Liability? {T}extual Analysis, Dictionaries, and 10-{K}s},
  journal = {The Journal of Finance},
  year    = {2011}, volume = {66}, number = {1}, pages = {35--65}}

@article{loughran2016,
  author  = {Loughran, Tim and McDonald, Bill},
  title   = {Textual Analysis in Accounting and Finance: A Survey},
  journal = {Journal of Accounting Research},
  year    = {2016}, volume = {54}, number = {4}, pages = {1187--1230}}

@inproceedings{hutto2014,
  author    = {Hutto, C. J. and Gilbert, Eric},
  title     = {{VADER}: A Parsimonious Rule-Based Model for Sentiment Analysis of Social Media Text},
  booktitle = {Proceedings of the Eighth International AAAI Conference on Weblogs and Social Media},
  year      = {2014}, pages = {216--225}}

@article{araci2019,
  author  = {Araci, Dogu},
  title   = {{FinBERT}: Financial Sentiment Analysis with Pre-trained Language Models},
  journal = {arXiv preprint arXiv:1908.10063},
  year    = {2019}}

@inproceedings{barbieri2020,
  author    = {Barbieri, Francesco and Camacho-Collados, Jose and Espinosa Anke, Luis and Neves, Leonardo},
  title     = {{TweetEval}: Unified Benchmark and Comparative Evaluation for Tweet Classification},
  booktitle = {Findings of the Association for Computational Linguistics: EMNLP 2020},
  year      = {2020}, pages = {1644--1650}}

@article{malo2014,
  author  = {Malo, Pekka and Sinha, Ankur and Korhonen, Pekka and Wallenius, Jyrki and Takala, Pyry},
  title   = {Good Debt or Bad Debt: Detecting Semantic Orientations in Economic Texts},
  journal = {Journal of the Association for Information Science and Technology},
  year    = {2014}, volume = {65}, number = {4}, pages = {782--796}}

@article{cronbach1955,
  author  = {Cronbach, Lee J. and Meehl, Paul E.},
  title   = {Construct Validity in Psychological Tests},
  journal = {Psychological Bulletin},
  year    = {1955}, volume = {52}, number = {4}, pages = {281--302}}

@inproceedings{chiu2016,
  author    = {Chiu, Billy and Korhonen, Anna and Pyysalo, Sampo},
  title     = {Intrinsic Evaluation of Word Vectors Fails to Predict Extrinsic Performance},
  booktitle = {Proceedings of the 1st Workshop on Evaluating Vector-Space Representations for NLP (RepEval)},
  year      = {2016}, pages = {1--6}}

@inproceedings{jacobs2021,
  author    = {Jacobs, Abigail Z. and Wallach, Hanna},
  title     = {Measurement and Fairness},
  booktitle = {Proceedings of the 2021 ACM Conference on Fairness, Accountability, and Transparency},
  year      = {2021}, pages = {375--385}}

@article{lopezlira2023,
  author  = {Lopez-Lira, Alejandro and Tang, Yuehua},
  title   = {Can {ChatGPT} Forecast Stock Price Movements? {R}eturn Predictability and Large Language Models},
  journal = {arXiv preprint arXiv:2304.07619},
  year    = {2023}}

@article{kirtac2024,
  author  = {Kirtac, Kemal and Germano, Guido},
  title   = {Sentiment Trading with Large Language Models},
  journal = {arXiv preprint arXiv:2412.19245},
  year    = {2024}}

@article{gilardi2023,
  author  = {Gilardi, Fabrizio and Alizadeh, Meysam and Kubli, Ma{\"e}l},
  title   = {{ChatGPT} Outperforms Crowd Workers for Text-Annotation Tasks},
  journal = {Proceedings of the National Academy of Sciences},
  year    = {2023}, volume = {120}, number = {30}, pages = {e2305016120}}

@article{glasserman2023,
  author  = {Glasserman, Paul and Lin, Caden},
  title   = {Assessing Look-Ahead Bias in Stock Return Predictions Generated by Large Language Models},
  journal = {arXiv preprint arXiv:2309.17322},
  year    = {2023}}

@article{he2025,
  author  = {He, Songrun and Lv, Linying and Manela, Asaf and Wu, Jimmy},
  title   = {Chronologically Consistent Large Language Models},
  journal = {arXiv preprint arXiv:2502.21206},
  year    = {2025}}

@article{bollen2011,
  author  = {Bollen, Johan and Mao, Huina and Zeng, Xiaojun},
  title   = {Twitter Mood Predicts the Stock Market},
  journal = {Journal of Computational Science},
  year    = {2011}, volume = {2}, number = {1}, pages = {1--8}}

@article{sprenger2014,
  author  = {Sprenger, Timm O. and Tumasjan, Andranik and Sandner, Philipp G. and Welpe, Isabell M.},
  title   = {Tweets and Trades: The Information Content of Stock Microblogs},
  journal = {European Financial Management},
  year    = {2014}, volume = {20}, number = {5}, pages = {926--957}}

@article{ranco2015,
  author  = {Ranco, Gabriele and Aleksovski, Darko and Caldarelli, Guido and Gr{\v{c}}ar, Miha and Mozeti{\v{c}}, Igor},
  title   = {The Effects of {T}witter Sentiment on Stock Price Returns},
  journal = {PLOS ONE},
  year    = {2015}, volume = {10}, number = {9}, pages = {e0138441}}

@article{cookson2020,
  author  = {Cookson, J. Anthony and Niessner, Marina},
  title   = {Why Don't We Agree? {E}vidence from a Social Network of Investors},
  journal = {The Journal of Finance},
  year    = {2020}, volume = {75}, number = {1}, pages = {173--228}}

@article{gande2009,
  author  = {Gande, Amar and Lewis, Craig M.},
  title   = {Shareholder-Initiated Class Action Lawsuits: Shareholder Wealth Effects and Industry Spillovers},
  journal = {Journal of Financial and Quantitative Analysis},
  year    = {2009}, volume = {44}, number = {4}, pages = {823--850}}

@article{kim2012,
  author  = {Kim, Irene and Skinner, Douglas J.},
  title   = {Measuring Securities Litigation Risk},
  journal = {Journal of Accounting and Economics},
  year    = {2012}, volume = {53}, number = {1--2}, pages = {290--310}}

@article{karpoff2008,
  author  = {Karpoff, Jonathan M. and Lee, D. Scott and Martin, Gerald S.},
  title   = {The Cost to Firms of Cooking the Books},
  journal = {Journal of Financial and Quantitative Analysis},
  year    = {2008}, volume = {43}, number = {3}, pages = {581--611}}

@article{miller2006,
  author  = {Miller, Gregory S.},
  title   = {The Press as a Watchdog for Accounting Fraud},
  journal = {Journal of Accounting Research},
  year    = {2006}, volume = {44}, number = {5}, pages = {1001--1033}}

@article{dyck2010,
  author  = {Dyck, Alexander and Morse, Adair and Zingales, Luigi},
  title   = {Who Blows the Whistle on Corporate Fraud?},
  journal = {The Journal of Finance},
  year    = {2010}, volume = {65}, number = {6}, pages = {2213--2253}}

@inproceedings{xie2023pixiu,
  author    = {Xie, Qianqian and Han, Weiguang and Zhang, Xiao and Lai, Yanzhao
               and Peng, Min and Lopez-Lira, Alejandro and Huang, Jimin},
  title     = {{PIXIU}: A Comprehensive Benchmark, Instruction Dataset and Large
               Language Model for Finance},
  booktitle = {Advances in Neural Information Processing Systems 36
               (Datasets and Benchmarks Track)},
  year      = {2023},
  url       = {https://proceedings.neurips.cc/paper_files/paper/2023/hash/6a386d703b50f1cf1f61ab02a15967bb-Abstract-Datasets_and_Benchmarks.html}
}

@inproceedings{xie2024finben,
  author    = {Xie, Qianqian and Han, Weiguang and Chen, Zhengyu and Xiang, Ruoyu
               and Zhang, Xiao and Ananiadou, Sophia and Huang, Jimin and others},
  title     = {{FinBen}: A Holistic Financial Benchmark for Large Language Models},
  booktitle = {Advances in Neural Information Processing Systems 37
               (Datasets and Benchmarks Track)},
  year      = {2024}
}

@article{cresci2019,
  author  = {Cresci, Stefano and Lillo, Fabrizio and Regoli, Daniele and
             Tardelli, Serena and Tesconi, Maurizio},
  title   = {Cashtag Piggybacking: Uncovering Spam and Bot Activity in Stock
             Microblogs on {Twitter}},
  journal = {ACM Transactions on the Web},
  volume  = {13},
  number  = {2},
  pages   = {1--27},
  year    = {2019},
  doi     = {10.1145/3313184}
}

@article{tetlock2011,
  author  = {Tetlock, Paul C.},
  title   = {All the News That's Fit to Reprint: Do Investors React to Stale
             Information?},
  journal = {The Review of Financial Studies},
  volume  = {24},
  number  = {5},
  pages   = {1481--1512},
  year    = {2011},
  doi     = {10.1093/rfs/hhq141}
}

@article{choi2024business,
  author  = {Choi, Stephen J. and Erickson, Jessica and Pritchard, A. C.},
  title   = {The Business of Securities Class Action Lawyering},
  journal = {Indiana Law Journal},
  volume  = {99},
  number  = {3},
  year    = {2024}
}

\appendix

\section{Relevance Filtering}
\label{app:filtering}

Table~\ref{tab:relevant-leadlag} repeats the lead--lag analysis of
Table~\ref{tab:leadlag} on messages the LLM instrument labels
\emph{relevant}, and Table~\ref{tab:relevance-decomp} reports the change
against the all-message baseline. Filtering shifts the same-day
correlation by at most $0.0164$ and never reverses a sign, so the central
result is not produced by the relevance labels. This matters because those
labels come from one of the instruments under evaluation; if the result
depended on them the comparison would be circular. At case level,
filtering reduces the common sample from 134 to 50 because windows
containing no relevant message become undefined, and every instrument
becomes nonsignificant.

\begin{table}[t]
\centering\small
\setlength{\tabcolsep}{2.5pt}
\begin{tabular}{@{}llrrrr@{}}
\toprule
Method & Score & $\rho_0$ & $\rho_1$ & $q_1$ & $n$ \\
\midrule
VADER   & coarse & $-0.0464^{***}$ & $-0.0195$        & 0.268 & 5{,}071 \\
        & graded & $-0.0417^{**}$  & $-0.0219$        & 0.194 & 5{,}502 \\
LM      & coarse & $-0.0293^{\dagger}$ & $-0.0160$    & 0.415 & 4{,}302 \\
        & graded & $-0.0425^{**}$  & $-0.0189$        & 0.306 & 4{,}372 \\
FinBERT & coarse & $-0.0834^{***}$ & $-0.0457^{**}$   & 0.023 & 3{,}247 \\
        & graded & $-0.0527^{***}$ & $-0.0259^{*}$    & 0.088 & 6{,}167 \\
RoBERTa & coarse & $-0.0835^{***}$ & $-0.0007$        & 0.972 & 2{,}511 \\
        & graded & $-0.0853^{***}$ & $-0.0350^{**}$   & 0.016 & 6{,}167 \\
Claude  & coarse & $-0.0866^{***}$ & $-0.0239$        & ---   & 4{,}279 \\
        & graded & $-0.0874^{***}$ & $-0.0336^{*}$    & ---   & 4{,}301 \\
\bottomrule
\end{tabular}
\caption{Lead--lag correlations restricted to relevant messages. LM denotes
Loughran--McDonald; RoBERTa denotes Twitter-RoBERTa. $q_1$ is FDR-adjusted
across the baseline family; the LLM reference is not included in that
correction (---). Significance as in Section~\ref{sec:setup}.}
\label{tab:relevant-leadlag}
\end{table}

\begin{table}[t]
\centering\small
\setlength{\tabcolsep}{4pt}
\begin{tabular}{@{}llrrr@{}}
\toprule
Method & Score & All & Relevant & $\Delta$ \\
\midrule
Claude  & coarse & $-0.0822$ & $-0.0866$ & $-0.0044$ \\
        & graded & $-0.0821$ & $-0.0874$ & $-0.0053$ \\
VADER   & coarse & $-0.0454$ & $-0.0464$ & $-0.0010$ \\
        & graded & $-0.0454$ & $-0.0417$ & $+0.0037$ \\
LM      & coarse & $-0.0249$ & $-0.0293$ & $-0.0044$ \\
        & graded & $-0.0559$ & $-0.0425$ & $+0.0134$ \\
FinBERT & coarse & $-0.0670$ & $-0.0834$ & $-0.0164$ \\
        & graded & $-0.0526$ & $-0.0527$ & $-0.0002$ \\
RoBERTa & coarse & $-0.0967$ & $-0.0835$ & $+0.0132$ \\
        & graded & $-0.0760$ & $-0.0853$ & $-0.0093$ \\
\bottomrule
\end{tabular}
\caption{Effect of relevance filtering on the same-day correlation.
$\Delta$ is relevant minus all.}
\label{tab:relevance-decomp}
\end{table}

\begin{table*}[t]
\centering\small
\begin{tabular}{@{}lllrrrrr@{}}
\toprule
 & & & \multicolumn{4}{c}{Forward Granger} & Distributed lag \\
\cmidrule(lr){4-7}\cmidrule(lr){8-8}
Method & Score & Sample & $F$ & $p$ & $q$ & $\beta_0$ & $q_\beta$ \\
\midrule
Claude  & coarse & all  & $5.435^{**}$          & 0.0051 & ---            & $-0.0086^{***}$ & --- \\
        &        & rel. & $7.888^{***}$         & 0.0005 & ---            & $-0.0089^{***}$ & --- \\
        & graded & all  & $3.909^{*}$           & 0.0218 & ---            & $-0.0071^{***}$ & --- \\
        &        & rel. & $4.717^{*}$           & 0.0101 & ---            & $-0.0071^{***}$ & --- \\
\addlinespace
VADER   & coarse & all  & $0.870$               & 0.4207 & 0.434          & $-0.0016^{*}$   & 0.050 \\
        &        & rel. & $2.406^{\dagger}$     & 0.0932 & 0.130          & $-0.0020^{\dagger}$ & 0.133 \\
        & graded & all  & $0.447$               & 0.6403 & 0.640          & $-0.0035^{*}$   & 0.069 \\
        &        & rel. & $0.879$               & 0.4169 & 0.434          & $-0.0038$       & 0.245 \\
\addlinespace
LM      & coarse & all  & $3.062^{*}$           & 0.0493 & 0.072          & $-0.0043^{***}$ & $<0.001$ \\
        &        & rel. & $4.388^{*}$           & 0.0139 & \textbf{0.034} & $-0.0067^{***}$ & $<0.001$ \\
        & graded & all  & $4.056^{*}$           & 0.0189 & \textbf{0.040} & $-0.0895^{***}$ & $<0.001$ \\
        &        & rel. & $3.265^{*}$           & 0.0406 & 0.065          & $-0.1140^{***}$ & $<0.001$ \\
\addlinespace
FinBERT & coarse & all  & $3.341^{*}$           & 0.0376 & 0.063          & $-0.0049^{**}$  & 0.003 \\
        &        & rel. & $3.907^{*}$           & 0.0219 & \textbf{0.044} & $-0.0084^{***}$ & $<0.001$ \\
        & graded & all  & $4.316^{*}$           & 0.0148 & \textbf{0.034} & $-0.0062^{***}$ & 0.001 \\
        &        & rel. & $3.466^{*}$           & 0.0335 & 0.059          & $-0.0102^{***}$ & 0.001 \\
\addlinespace
RoBERTa & coarse & all  & $2.111$               & 0.1242 & 0.159          & $-0.0088^{***}$ & $<0.001$ \\
        &        & rel. & $0.944$               & 0.3910 & 0.431          & $-0.0106^{***}$ & $<0.001$ \\
        & graded & all  & $1.589$               & 0.2069 & 0.245          & $-0.0144^{***}$ & $<0.001$ \\
        &        & rel. & $1.534$               & 0.2185 & 0.250          & $-0.0166^{***}$ & $<0.001$ \\
\bottomrule
\end{tabular}
\caption{Forward-Granger and distributed-lag results. Bold $q$ marks the four
specifications that survive FDR correction. All-message rows use
$n_F = 103{,}184$ and $n_\beta = 103{,}005$; relevant-message rows use
$99{,}328$ and $99{,}157$. Coefficients are not comparable across instruments
(see text). Significance as in Section~\ref{sec:setup}.}
\label{tab:granger-full}
\end{table*}

\section{Forward-Granger and Distributed-Lag Tests}
\label{app:granger}

Table~\ref{tab:granger-full} gives the complete battery summarised in
Section~\ref{sec:timing}. Distributed-lag coefficients are not comparable
across instruments because the underlying signals differ in scale; only
sign and significance transfer.

\section{Common Day Intersection, Relevant Messages}
\label{app:common}

Table~\ref{tab:common} in the main text holds $n$ fixed across instruments
using all messages. Table~\ref{tab:common-rel} repeats that intersection
on relevant messages only. The reordering described in
Section~\ref{sec:instability} persists: FinBERT is strongest same day on
the coarse panel and Twitter-RoBERTa on the graded panel, while VADER
remains indistinguishable from zero throughout.

\begin{table}[t]
\centering\small
\setlength{\tabcolsep}{4pt}
\begin{tabular}{@{}llrrr@{}}
\toprule
Panel & Method & $\rho_0$ & $\rho_1$ & $\rho_2$ \\
\midrule
C/rel. & Claude  & $-0.0910^{***}$ & $-0.0395$ & $-0.0288$ \\
       & VADER   & $+0.0084$ & $-0.0043$ & $+0.0442$ \\
       & LM      & $-0.0564^{*}$ & $-0.0467^{\dagger}$ & $-0.0446$ \\
       & FinBERT & $-0.1417^{***}$ & $-0.0357$ & $-0.0153$ \\
       & RoBERTa & $-0.0517^{\dagger}$ & $+0.0105$ & $-0.0208$ \\
\midrule
G/rel. & Claude  & $-0.0711^{***}$ & $-0.0298^{\dagger}$ & $-0.0514^{**}$ \\
       & VADER   & $-0.0216$ & $-0.0098$ & $+0.0109$ \\
       & LM      & $-0.0357^{*}$ & $-0.0154$ & $-0.0139$ \\
       & FinBERT & $-0.0295^{\dagger}$ & $-0.0144$ & $+0.0206$ \\
       & RoBERTa & $-0.0706^{***}$ & $-0.0368^{*}$ & $-0.0262$ \\
\bottomrule
\end{tabular}
\caption{Common case day intersection on relevant messages. C and G
denote coarse and graded; $n = 1{,}328$ and $3{,}295$ respectively.}
\label{tab:common-rel}
\end{table}

\section{Look Ahead Contamination}
\label{app:contamination}

Section~\ref{sec:robustness} reports the headline contamination test. This
appendix gives the decomposition. Table~\ref{tab:contam} reports the
difference in differences against each baseline separately rather than
against their mean, so the null is not an artifact of averaging: all
sixteen estimates lie between $-0.0400$ and $+0.0281$ and every interval
spans zero.

Within the archive cohort, splitting at the median case year gives an LLM
advantage of $+0.0395$ for older cases against $+0.0363$ for newer ones on
the coarse same day measure, and $+0.0153$ against $+0.0127$ at a one day
lag. The advantage is slightly larger for older cases, but the differences are small. This comparison alone does not establish contamination.

We are more cautious about the case level measure. Splitting that test by
cohort leaves 80 archive and 47 recent cases, and on a reconstruction of
the panel the LLM's advantage is larger in the archive than in the recent
cohorts, which is the direction contamination predicts. The subsample is
small and we do not treat this as evidence of contamination, but the
daily panel null does not extend to it, which is why the paper rests its
claims on the daily panel.

\begin{table}[t]
\centering\small
\setlength{\tabcolsep}{3pt}
\begin{tabular}{@{}llrr@{}}
\toprule
Repr. & vs baseline & DiD ($\ell=0$) & DiD ($\ell=1$) \\
\midrule
coarse & FinBERT & $+0.0175$ & $+0.0281$ \\
       & Loughran--McDonald & $-0.0170$ & $-0.0400$ \\
       & Twitter-RoBERTa & $+0.0035$ & $+0.0078$ \\
       & VADER & $+0.0202$ & $+0.0111$ \\
\midrule
graded & FinBERT & $+0.0044$ & $+0.0134$ \\
       & Loughran--McDonald & $-0.0156$ & $-0.0277$ \\
       & Twitter-RoBERTa & $+0.0221$ & $+0.0066$ \\
       & VADER & $+0.0254$ & $+0.0026$ \\
\bottomrule
\end{tabular}
\caption{Per baseline difference in differences. Each cell is the LLM's
$|\rho|$ advantage over that baseline in the archive cohort minus the same
advantage in the 2024--2025 cohorts. No estimate is distinguishable from
zero at 95\% over 1{,}000 case level bootstrap resamples.}
\label{tab:contam}
\end{table}

\FloatBarrier
\section{Retained Null Results}
\label{app:nulls}

For completeness: the fear/panic/outrage \emph{share} is insignificant at
every lag, while its daily count is not; several event window
specifications lose significance once firm and sector controls are added;
relevance filtering does not create the daily result; the filtered
case level measure becomes sparse and nonsignificant; negative horizon
local projections limit causal interpretation and total, relevant, and
negative message volume all fail to predict settlement size or case level
price damage depth.

\section{Event Validation and Local Projections}
\label{app:event}
Mean negative message share rises from 11.8\% in the baseline window to
46.7\% in the disclosure window ($n = 184$, paired $t = 10.97$,
$p < 10^{-21}$), confirming that the complaint derived disclosure dates
coincide with an information shock rather than background chatter. With
controls for log market capitalisation, class period length, defendant
count, and sector, class period negative share predicts maximum drawdown
($n = 41$, $t = -3.24$, $p = 0.003$, adjusted $R^2 = 0.60$).

\begin{figure}[t]
\centering
\includegraphics[width=.95\columnwidth]{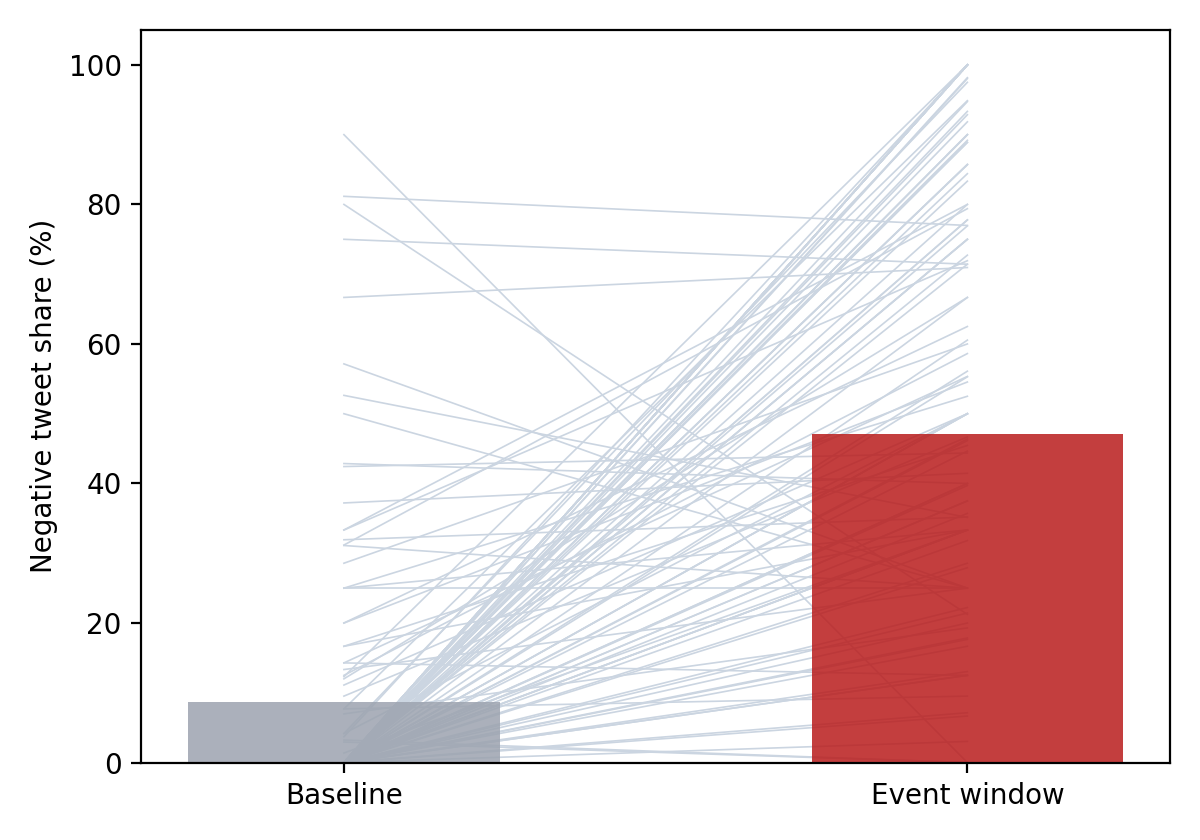}

\includegraphics[width=.95\columnwidth]{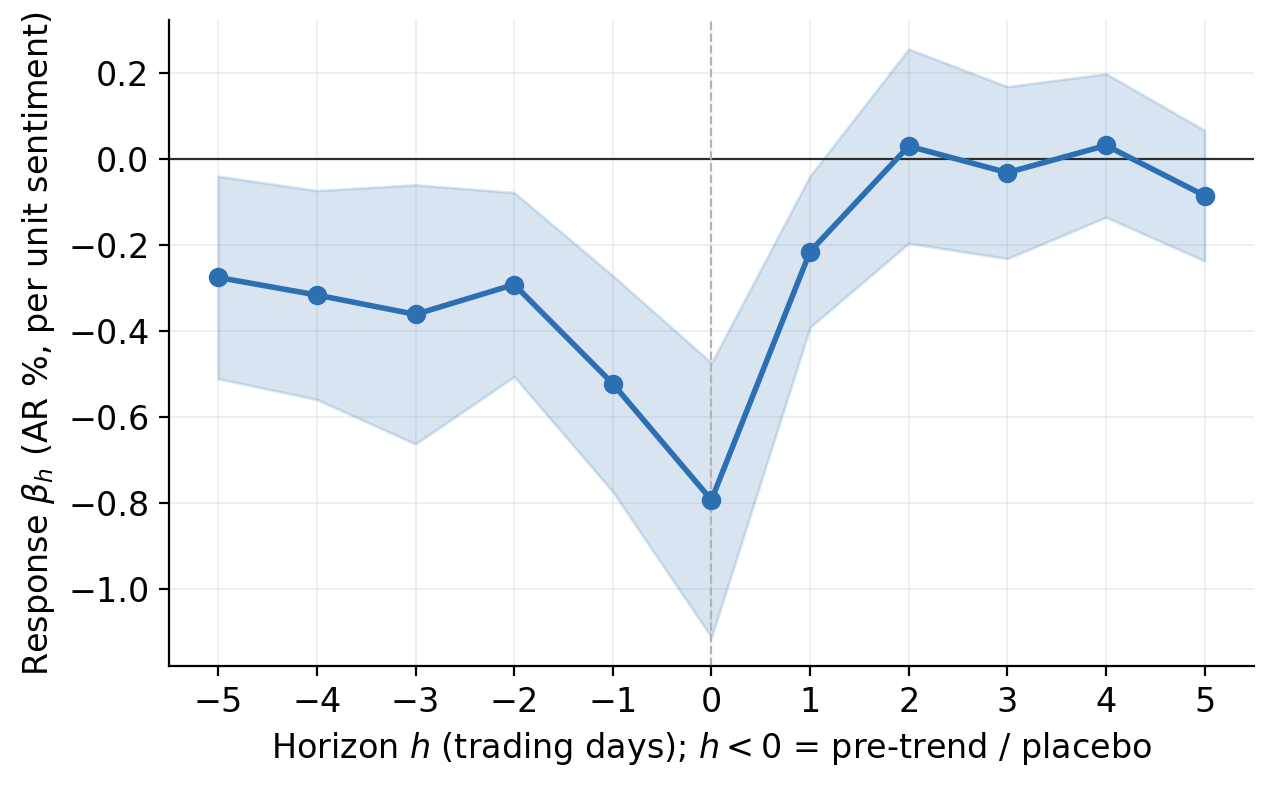}
\caption{Top: the complaint derived disclosure window is a sharp negativity shock. Bottom: local projection coefficients peak contemporaneously and decay after one day; the significantly negative coefficients at negative horizons are a failed placebo and caution against a causal interpretation.}
\label{fig:lp}
\end{figure}

Local projections (Figure~\ref{fig:lp}, bottom) peak at $h=0$
($\beta = -0.0079$, $p<0.001$), remain significant at $h=1$
($\beta = -0.0022$, $p<0.05$), and are indistinguishable from zero by
$h=2$. Coefficients at negative horizons are also significantly negative,
and a reverse next-day correlation is present ($\rho = -0.046$,
$p<0.01$). 

\section{Rank Correlations Between the Two Validities}
\label{app:rankcorr}

\begin{figure*}[t]
\centering
\includegraphics[width=\textwidth]{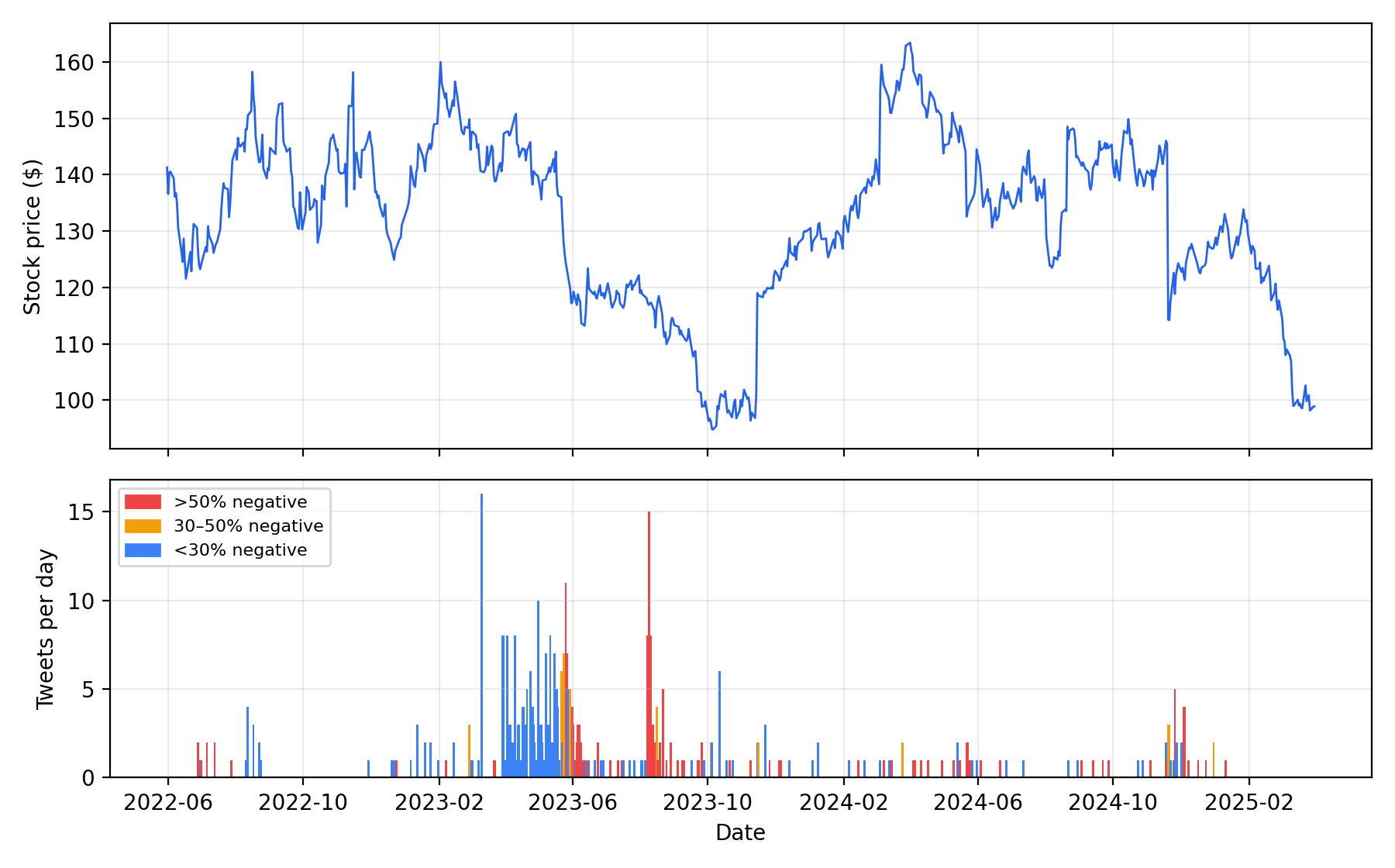}
\caption{Target Corporation over its class period. Top: daily closing price. Bottom: messages per day, coloured by the share of negative messages. Negative-heavy days concentrate around the major price declines.}
\label{fig:tgt}
\end{figure*}

Table~\ref{tab:rankcorr} gives every correlation between a construct validity
measure and a predictive validity column, completing Table~\ref{tab:joint}.
With five instruments the asymptotic $p$-values reported by standard
software are unreliable, so we compute exact two sided $p$-values by
enumerating all $5! = 120$ permutations.

\begin{table}[ht]
\centering\small
\begin{tabular}{@{}lrrrr@{}}
\toprule
& \multicolumn{2}{c}{coarse} & \multicolumn{2}{c}{graded} \\
\cmidrule(lr){2-3}\cmidrule(lr){4-5}
Measure & $|\rho_0|$ & $|\rho_1|$ & $|\rho_0|$ & $|\rho_1|$ \\
\midrule
$\kappa$   & $+0.50$ & $-0.30$ & $+0.90$ & $+0.40$ \\
           & (0.450) & (0.683) & (0.083) & (0.517) \\
Accuracy   & $+0.60$ & $-0.10$ & $+1.00$ & $+0.70$ \\
           & (0.350) & (0.950) & (0.017) & (0.233) \\
Macro-F1   & $+0.60$ & $-0.10$ & $+1.00$ & $+0.70$ \\
           & (0.350) & (0.950) & (0.017) & (0.233) \\
\bottomrule
\end{tabular}
\caption{Spearman rank correlation between each construct validity measure
and each predictive validity column, over the five instruments. Exact
two sided $p$-values from all 120 permutations in parentheses. Accuracy
and macro-F1 give identical values because they order the five instruments
identically. Only the graded same day column is nominally significant, and
no cell survives adjustment for twelve comparisons.}
\label{tab:rankcorr}
\end{table}

Across all three agreement measures, the closest alignment is with graded same-day associations. At one day, the relationship is weaker for graded scores and weakly negative for coarse scores. None of the twelve comparisons survives multiple-testing correction. With only five instruments, we treat these patterns as descriptive.

\section{Case Study: Target Corporation}
\label{app:tgt}

A large cap example shows the mechanism: Figure~\ref{fig:tgt} overlays Target Corporation's stock price with
its daily message volume coloured by negative share over the class period.
Days on which more than half the messages are negative cluster around the
two largest price declines, while low negativity days dominate the calm
stretches and carry little price information. Message \emph{volume} is
similar in negative and quiet periods; it is the classified sentiment that
lines up with the moves. The case illustrates at the single case level the
domain result of Section~\ref{sec:predictive}.
\end{document}